\documentclass[runningheads]{llncs}
\usepackage{graphicx}
\usepackage{amsmath,amssymb} 
\usepackage{color}

\begin{document}
\newcommand\eg{{\it e.g.}}

\title{Modeling Visual Context is Key to \\ Augmenting Object Detection Datasets} 
\titlerunning{Context Data Augmentation for Object Detection}
\author{Nikita Dvornik, Julien Mairal, Cordelia Schmid}
\authorrunning{N. Dvornik, J. Mairal and C. Schmid}
\institute{Inria}


\institute{Univ. Grenoble Alpes, Inria, CNRS, Grenoble INP\thanks{Institute
of Engineering Univ. Grenoble Alpes}, LJK, 38000
Grenoble, France\\
  \email{ firstname.lastname@inria.fr}
}

\maketitle

\begin{abstract}
   Performing data augmentation for learning deep neural networks is well known to be 
important for training visual recognition
systems. By artificially increasing the number of training examples, it helps
reducing overfitting and improves generalization. For object detection,
classical approaches for data augmentation consist of generating images
obtained by basic geometrical transformations and color changes of original
training images. In this work, we go one step further and leverage segmentation
annotations to increase the number of object instances present on training
data. For this approach to be successful, we show that modeling appropriately
the visual context surrounding objects is crucial to place them in the
right environment. Otherwise, we show that the previous
strategy actually hurts. With our context model, we achieve significant mean
average precision improvements when few labeled examples are
available on the VOC'12 benchmark.


\end{abstract}

\section{Introduction}
\begin{figure}[hbtp!]
\begin{center}
   \includegraphics[width=0.99\linewidth,trim=0 60 0 0,clip]{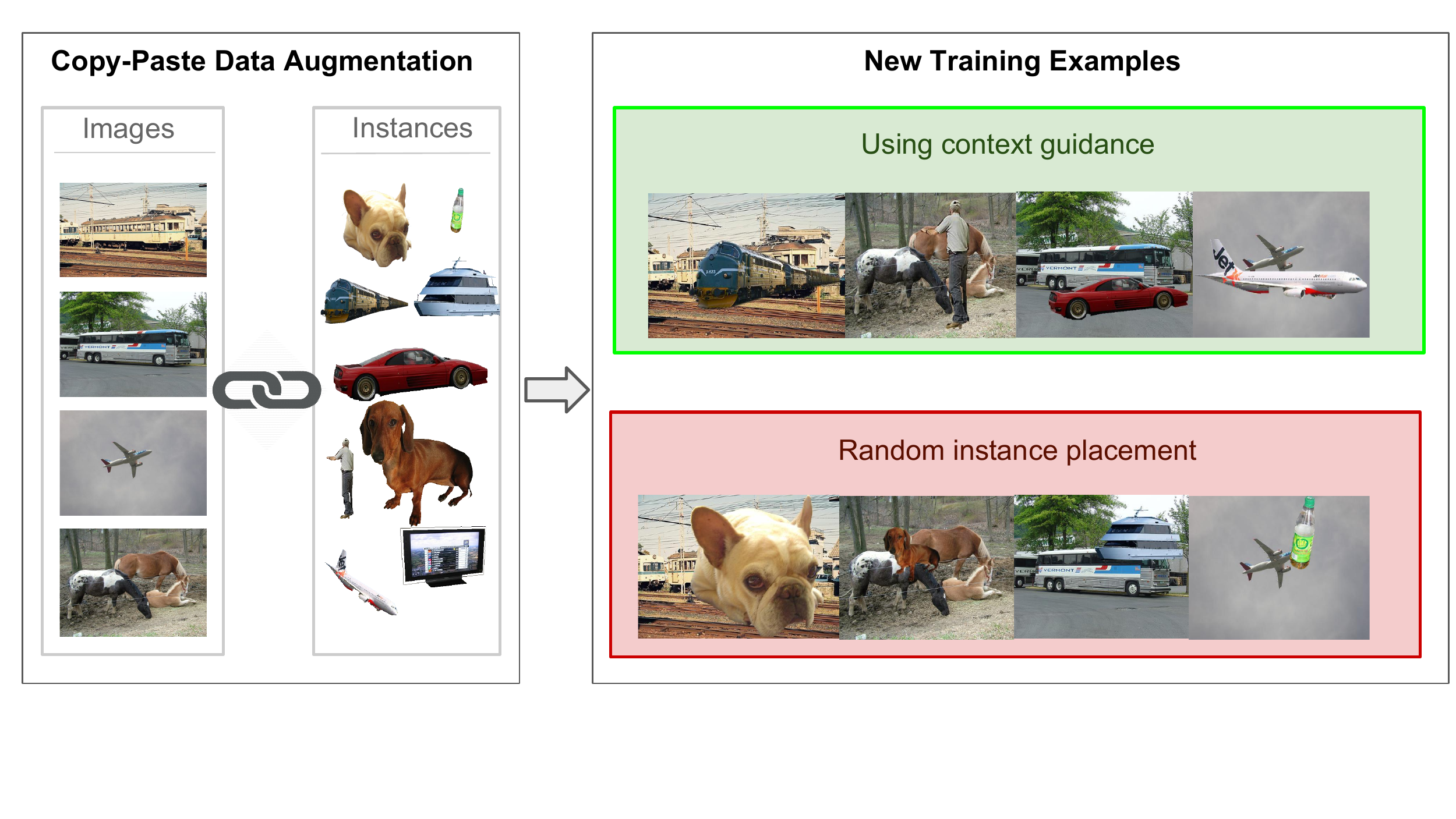}
\end{center}
\vspace*{-0.3cm}
\caption{\textbf{Examples of data-augmented training examples produced by our approach.}
Images and objects are taken from the VOC'12 dataset that contains segmentation annotations.
We compare the output obtained by pasting the objects with our context model vs. those obtained with random placements.
Even though the results are not perfectly photorealistic and display blending
artefacts, the visual context surrounding objects is more often correct with the explicit context model.
 }
\label{fig:segmentations2}
\vspace*{-0.4cm}
\end{figure}

Object detection is one of the most classical computer vision task and
is often considered as a basic proxy for scene understanding.  Given an input image, an
algorithm is expected to produce a set of tight boxes around objects while
automatically classifying them.  Obviously, modeling correctly object appearances is
important, but it is also well-known that visual context provides important
cues for recognition, both for computer vision systems and for
humans~\cite{oliva2007role}.

Objects from the same class tend indeed to be grouped together in similar
environments; sometimes they interact with it and do not even make sense
in its absence. Whenever visual information is corrupted, ambiguous, or incomplete 
(\eg, an image contains noise, bad illumination conditions, or an object is
occluded or truncated), visual context becomes a crucial source of information.
Frequently, certain object categories may for instance most often appear in specific
conditions (\eg, planes in the sky, plates on the table), in co-occurrence with
objects of other specific classes (\eg, baseball ball and baseball bat), and more 
generally, any type of clue for object recognition that is not directly related to
the object's appearance is named ``context'' in the literature. 
For this reason, a taxonomy of contextual information is
proposed in \cite{divvala2009empirical} to better understand what type of visual
context is useful for object detection.

Before the deep learning/ImageNet revolution, the previous generation of object
detectors such as
\cite{murphy2006object,felzenszwalb2010object,park2010multiresolution,heitz2008learning}
modeled the interaction between object locations, categories, and context by
manual engineering of local descriptors, feature aggregation methods, and by
defining structural relationship between objects.
In contrast, recent works based on convolutional neural networks such
as~\cite{faster-rcnn,fast-rcnn,ssd,yolo} implicitly model visual context by
design since the receptive field of ``artificial neurons'' grows with the
network's depth, eventually covering the full image for the last layers.  For
this reason, these CNNs-based  approaches have shown modest improvements when
combined with an explicit context model~\cite{yu2016role}.

Our results are not in contradiction with such previous findings. We show that
explicit context modeling is important only for a particular part of object
detection pipelines that was not considered in previous work. When
training a convolutional neural network, it is indeed important to control
overfitting, especially if few labeled training examples are
available. Various heuristics are typically used for that purpose such as
DropOut~\cite{srivastava2014dropout}, penalizing the norm of the network
parameters (also called weight decay), or early stopping the optimization
algorithm. Even though the exact regularization effect of such approaches on
learning is not well understood from a theoretical point of view,  
these heuristics have been found to be useful in practice.

Besides these heuristics related to the learning procedure, another way to
control overfitting consists of artificially increasing the size of training
data by using prior knowledge on the task. For instance, all object classes
from the VOC'12 dataset~\cite{pascal} are invariant to horizontal
flips (\eg, a flipped car is still a car) and to many less-trivial
transformations.
A more ambitious data augmentation technique consists of leveraging
segmentation annotations, either obtained manually, or from an automatic
segmentation system, and create new images with objects placed at various
positions in existing
scenes~\cite{cut_paste,gupta2016synthetic,georgakis2017synthesizing}. While
not achieving perfect photorealism, this strategy with random placements has
proven to be surprisingly effective for \emph{object instance
detection}~\cite{cut_paste}, which is a fine-grained detection task consisting
of retrieving instances of a particular object from an image collection; in
contrast, \emph{object detection} focuses on detecting object instances from a
particular category. Unfortunately, the random-placement strategy does not extend to
the object detection task, as shown in the experimental section. By placing
training objects at unrealistic positions, implicitely modeling context becomes difficult and
the detection accuracy drops substantially.

Along the same lines, the authors of~\cite{gupta2016synthetic} have proposed
to augment datasets for text recognition by adding text on images in a realistic
fashion. There, placing text with the right geometrical context proves to
be critical. Significant improvements in accuracy are obtained by first
estimating the geometry of the scene, before placing text on an estimated plane.
Also related, the work of \cite{georgakis2017synthesizing} is using 
successfully such a data augmentation technique for object detection in indoor scene
environments. Modeling context has been found to be critical as well and has been achieved
by also estimating plane geometry and objects are typically placed on detected
tables or counters, which often occur in indoor scenes.

In this paper, we consider the general object detection problem, which requires
more generic context modeling than estimating plane and surfaces as done for instance
in~\cite{gupta2016synthetic,georgakis2017synthesizing}.
To this end, the first contribution of our paper is methodological:
we propose a context model based on a convolutional neural network,
which will be made available as an open-source software package.
The model estimates the likelihood of a particular category of object to be present
inside a box given its neighborhood, and then automatically finds suitable
locations on images to place new objects and perform data augmentation.
A brief illustration of the output produced by this approach is presented in Figure~\ref{fig:segmentations2}.
The second contribution is experimental: We show with extensive tests on
the VOC'12 benchmark that context modeling is in fact a key to obtain good 
results for object detection and that substantial improvements over non-data-augmented baselines may be achieved when few labeled
examples are available.

\section{Related Work}
In this section, we briefly discuss related work for visual context modeling
and data augmentation for object detection.

\paragraph{Modeling visual context for object detection.}
Relatively early, visual context has been modeled by computing statistical
correlation between low-level features of the global scene and descriptors
representing an object~\cite{torralba2001statistical,torralba2003contextual}.
Later, the authors of~\cite{felzenszwalb2010object} introduced a
simple context re-scoring approach operating on appearance-based detections. 
To encode more structure, graphical models
were then widely used in order to jointly model appearance, geometry, and
contextual relations~\cite{choi2010exploiting,gould2009decomposing}. Then, deep
learning approaches such as convolutional neural networks started to be used~\cite{faster-rcnn,fast-rcnn,ssd};
as mentioned previously, their features 
already contain implicitly contextual information. 
Yet, the work of~\cite{chu2018deep} explicitly incorporates higher-level context clues and
combines a conditional random field model with detections obtained by Faster-RCNN. 
With a similar goal, recurrent neural networks are used in \cite{bell2015inside} to model spatial locations
of discovered objects.
Another complementary direction in context modeling with
convolutional neural networks use a deconvolution pipeline
that increases the field of view of neurons and fuse features at different
scales~\cite{bell2015inside,blitznet,dssd}, showing better performance essentially
on small objects.
The works of \cite{divvala2009empirical,barnea2017utility} analyze
different types of contextual relationships, identifying the most useful ones for detection, as
well as various ways to leverage them. However, despite these efforts, an improvement
due to purely contextual information has always been relatively modest
\cite{yu2016role,yao2010modeling}.

\paragraph{Data augmentation for object detection.}
Data augmentation is a major tool to train deep neural networks. If varies from trivial
geometrical transformations such as horizontal flipping, cropping with color
perturbations, and adding noise to an image \cite{imagenet}, to synthesizing new
training images \cite{frid2018synthetic,peng2015learning}.
Some recent object detectors \cite{ssd,yolo,blitznet} benefit from standard data
augmentation techniques more than others \cite{faster-rcnn,fast-rcnn}.
The performance of Fast- and Faster-RCNN could be for instance increased by
simply corrupting random parts of an image in order to mimic
occlusions~\cite{random_erase}.
Regarding image synthesis, recent
works such as~\cite{karsch2011rendering,movshovitz2016useful,su2015render} build
and train their models on purely synthetic rendered 2d and 3d scenes. However, a major
difficulty for models
trained on synthetic images is to guarantee that they will generalize well to real data
since the synthesis process introduces significant changes of image statistics \cite{peng2015learning}. To address
this issue, the authors of \cite{gupta2016synthetic} adopt a different
direction by pasting real segmented object into natural images, which reduces
the presence of rendering artefacts. For object instance
detection, the
work~\cite{georgakis2017synthesizing} estimates scene geometry and spatial layout, before
synthetically placing objects in the image to create realistic training
examples. In \cite{cut_paste}, the authors propose an even simpler solution to
the same problem by pasting images in random positions but modeling well
occluded and truncated objects, and making the training step robust to boundary
artifacts at pasted locations.

\section{Modeling Visual Context for Data Augmentation}
\begin{figure}[t!]
\begin{center}
  \includegraphics[width=0.99\linewidth,trim=0 50 0 0,clip]{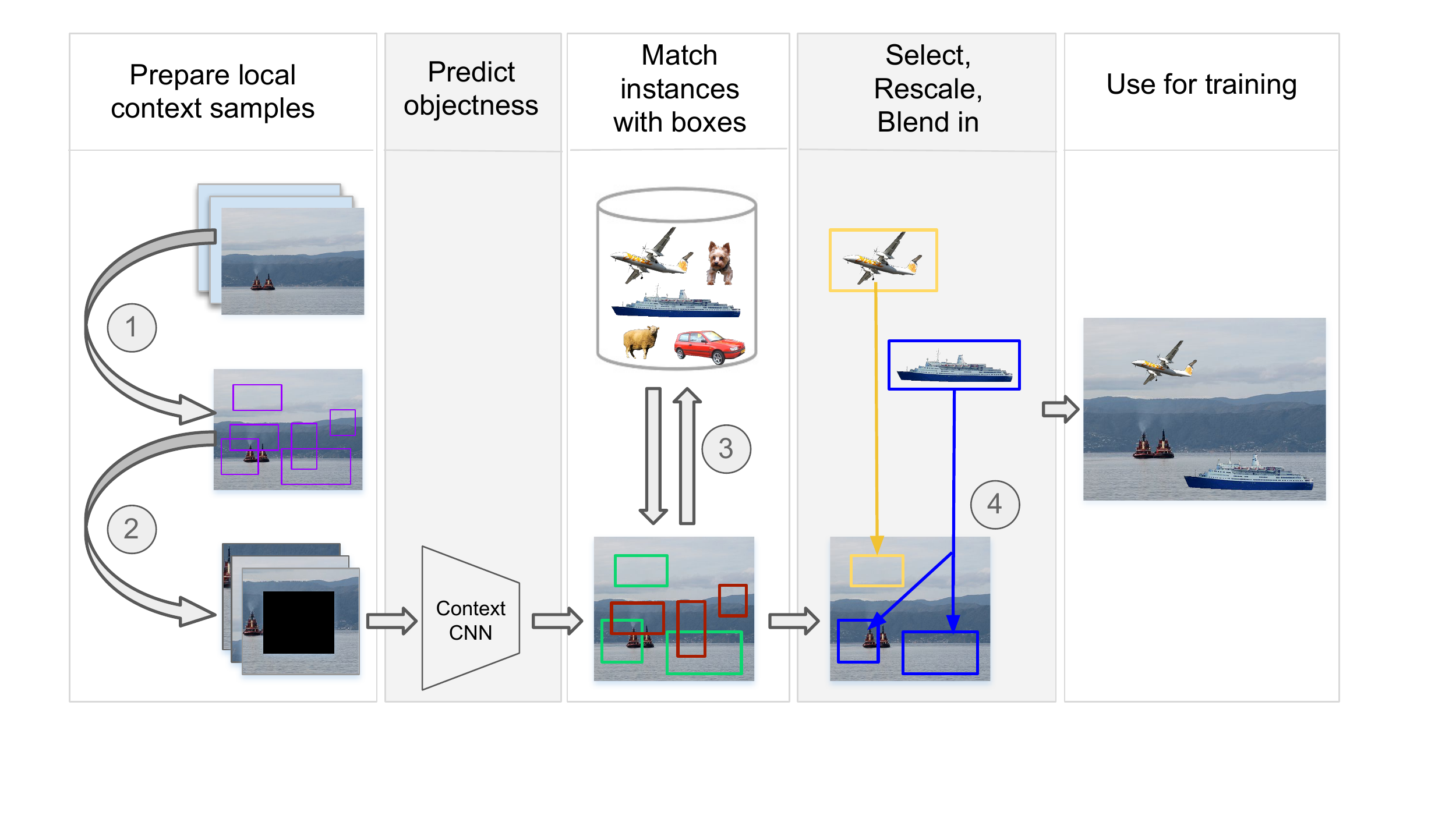}
\end{center}
\vspace{-0.5cm}
\caption{\textbf{Illustration of our data augmentation approach.}
  We select an image for augmentation and 1) generate 200 candidate boxes that cover
  the image. Then, 2) for each box we find a neighborhood that contains the box
  entirely, crop this neighborhood and mask all pixels falling inside
  the bounding box; this ``neighborhood'' with masked pixels is then fed to the context neural network module and 3)
  object instances are matched to boxes that have high confidence scores for the presence of an object category.
  4) We select at most two instances that are rescaled and blended into the selected bounding boxes.
  The resulting image is then used for training the object detector.}
\label{fig:diagram}
\vspace*{-0.5cm}
\end{figure}

Our approach for data augmentation mainly consists of two parts: we first model
visual context by using bounding box annotations, where the surrounding of a
box is used as an input to a convolutional neural network to predict the
presence or absence of an object within the box.  Then, the trained context
model is used to generate a set of possible new locations for objects. The full
pipeline is presented in Fig.~\ref{fig:diagram}.
In this section, we describe these two steps in details, but before that, we
present and discuss a preliminary experiment that has motivated our work.

\subsection{Preliminary Experiment with Random Positioning}\label{subsec:prelim}
In \cite{cut_paste}, data augmentation is performed by placing segmented objects 
at random positions in new scenes. As mentioned previously, the strategy
was shown to be effective for object instance detection, as soon as an appropriate
procedure is used for preventing the object detector to overfit blending
artefacts---that is, the main difficulty is to prevent the detector to ``detect
artefacts'' instead of detecting objects of interest.
This is achieved by using various blending strategies to smooth object boundaries such as Poisson
blending~\cite{perez2003poisson}, and by adding ``distractors'' that are
objects that do not belong to any of the dataset categories, but which are also
synthetically pasted on random backgrounds. With distractors, artefacts occur
both in positive and negative examples, for each of the categories, preventing
the network trained for object detection to overfit them.
According to~\cite{cut_paste}, this strategy can bring substantial improvements
for the object instance detection/retrieval task, where modeling the fine-grain
appearance of an object instance seems to be more important than modeling
visual context as in the general category object detection task.

Unfortunately, the above context-free strategy does not extend trivially to the
object detection task we consider. Our preliminary experiment conducted on the
VOC'12 dataset actually shows that it may even hurt the accuracy of the
detector, which has motivated us to propose instead an explicit context model.
Specifically, we conducted an experiment by following the original strategy
of~\cite{cut_paste} as closely as possible.  We use 
the subset of the VOC'12 train set that has 
ground-truth segmentation annotations 
to cut object instances from images and then place them
on other images from the training set. As in \cite{cut_paste}, we experimented with
various blending strategies (Gaussian or linear blur, Poisson blending, or
using no blending at all) to smooth the boundary artifacts.
Following~\cite{cut_paste}, we also considered ``distractors'', which are then
labeled as background. Distractors were simply obtained by copy-pasting
segmented objects from the COCO dataset~\cite{coco} from categories that do
not appear in VOC'12.\footnote{Note that external data from COCO was used only in this
preliminary experiment and not in the experiments reported later in Section~\ref{sec:exp}.}

For any combination of blending strategy, by using distractors or not, the naive
data augmentation approach with random placement did not improve upon the
baseline without data augmentation for the classical object detection task. 
A possible explanation may be that for instance object detection, the detector does not need to learn
intra-class variability of object/scene representations and seems to concentrate only on appearance
modeling of specific instances, which is not the case for category-level object detection.
This experiment was the key motivation for proposing a context model, which we now present.

\subsection{Modeling Visual Context with Convolutional Neural Networks}
Since the context-free data augmentation failed, we propose
to learn where to automatically place objects by using a convolutional neural
network. Here, we present the data generation, model training, and object
placement procedures.

\begin{figure}[hbtp!]
\begin{center}
  \includegraphics[width=0.99\linewidth,trim=90 115 230 30,clip]{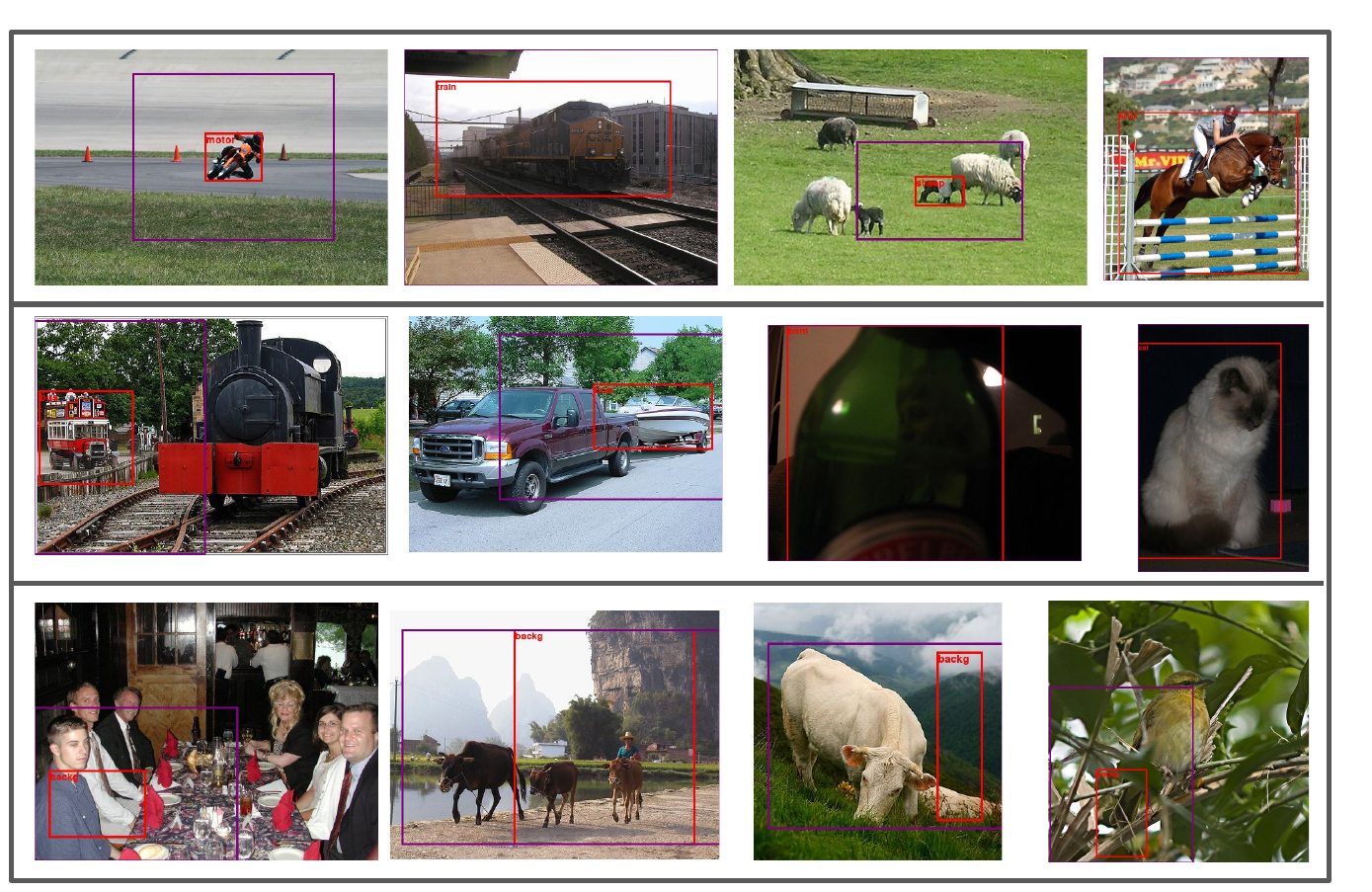}
\end{center}
\vspace{-0.5cm}
\caption{\textbf{Contextual images - examples of inputs to the context model}.
  A subimage bounded by a magenta box is used as an input to the context model
  after masking-out the object information inside a red box.
  The top row lists examples of positive samples encoding 
  real objects surrounded by regular and predictable context.
  Positive training examples with ambiguous or uninformative context are
  given in the second row.
  The bottom row depicts negative examples enclosing background.
  This figure shows that contextual images could be ambiguous to classify
  correctly and the task of predicting the category given only the context is
  challenging.}
\label{fig:context_input}
\end{figure}

\paragraph{Contextual data generation.} \label{sec:context_inputs}
We consider training data with bounding box and category
annotations. For each bounding box $B$ associated to a training image~$I$, we create a set
of training contexts, which are defined as subimages of $I$ fully enclosing the
bounding box~$B$ whose content is masked out, as illustrated in
Figure~\ref{fig:context_input}. Several contexts can be created from a single
annotated bounding box~$B$ by varying the size of the subimage around~$B$ and
its aspect ratio. In addition, ``background'' contexts are also created by
considering random bounding boxes whose intersection over union with any ground
truth doesn't exceed a threshold of $0.3$, and whose content is also masked out.
The shape of such boxes is defined by aspect ratio $a$ and relative scale $s$.
We draw a pair of parameters from the joint distribution induced by bounding
boxes containing positive objects, i.e. a $30 \times 30$ bins normalized 
histogram. Since in general, there is more background samples than the
ones actually containing objects, we sample ``background'' contexts $3$ times
more often following sampling strategies in \cite{faster-rcnn,ssd}.

\begin{figure}[hbtp!]
\begin{center}
  \includegraphics[width=0.99\linewidth,trim=100 0 250 300,clip]{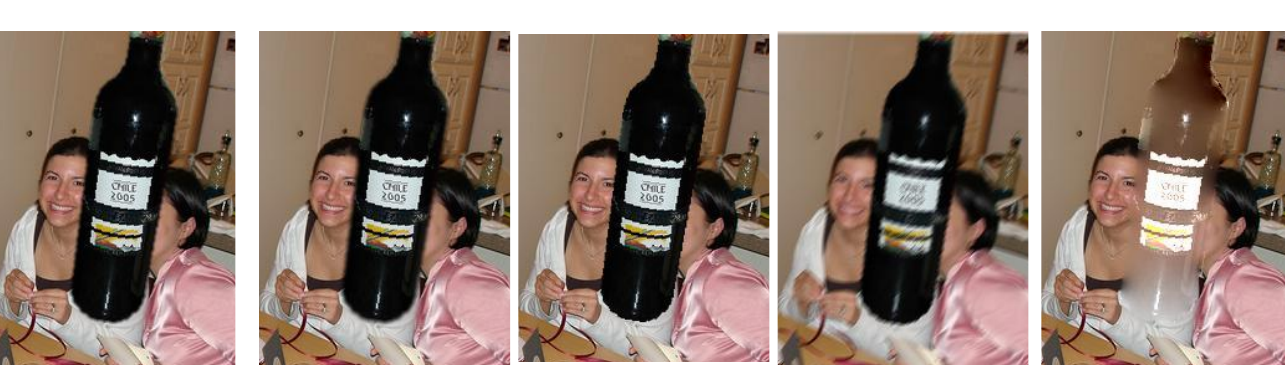}
\end{center}
\vspace{-0.5cm}
\caption{\textbf{Different kinds of blending used in experiments.}
  From left to right: linear smoothing of boundaries, Gaussian smoothing, no
  processing, motion blur of the whole image, Poisson blending \cite{perez2003poisson}.}
\label{fig:blendings}
\end{figure}

\paragraph{Model training.}
Given the set of all contexts, gathered from all training data, we train a
convolutional neural network to predict the presence of each object in the
masked bounding box. The input to the network are the ``contextual images''
obtained during the data generation step, and which contain a masked bounding
box inside. These contextual images are resized to $300 \times 300$ pixels, and
the output of the network is a label in a set $\{1,2,...,K+1\}$, where $K$ is the
number of object categories and the $(K+1)$-th class represents background. For
such a multi-class image classification problem here, we use the classical ResNet50
network~\cite{resnet} pre-trained on ImageNet, and change the last layer to be a
softmax with $K+1$ activations (see experimental section for details).

\paragraph{Selection of object locations at test time.}
Once the context model is trained by using training data annotated with
bounding boxes, we use it to select locations to perform data augmentation on a
given image. As input, the trained classifier receives ``contextual images''
with a bounding box masked out (as in Section~\ref{sec:context_inputs}). The
model is able to provide a set of ``probabilities'' representing the presence of
each object category in a given bounding box, by considering its visual
surrounding. Since evaluating all potential bounding boxes from an image is too
costly, we randomly draw 200 candidate bounding boxes and retain the ones where
an object category has a score greater than $0.8$; empirically, the number 200
was found to provide good enough bounding boxes among the top scoring ones,
while resulting in a reasonably fast data augmentation procedure.

\paragraph{Blending objects in their environment.}
Whenever a bounding box is selected by the previous procedure, we need to blend 
an object at the corresponding location. This step follows closely the findings
of~\cite{cut_paste}. We consider different types of blending techniques
(Gaussian or linear blur, simple copy-pasting with no post-processing, or generating
blur on the whole image to imitate motion), 
and randomly choose one of them in order to introduce a larger
diversity of blending artefacts.  We also do not consider Poisson blending in our approach,
which was considerably slowing down the data generation procedure.
Unlike~\cite{cut_paste} and unlike our
preliminary experiment described in Section~\ref{subsec:prelim}, 
we do not use distractors, which were found to be less important for our task
than in~\cite{cut_paste}.
As a consequence, we do not need to exploit external data to perform data
augmentation.  
Qualitative results are illustrated on Figure~\ref{fig:blendings}.

\section{Experiments}\label{sec:exp}
In this section, we present experiments demonstrating the importance of context
modeling for data augmentation. We evaluate our approach on the subset of the 
VOC'12 dataset that contains segmentation annotations, and study the impact of 
data augmentation when changing the amount of training data available.
In Section~\ref{sec:datasets}, we present data, tools, and evaluation metrics.
In Section~\ref{sec:details}, we present 
implementation details that are common
to all experiments, in order to make our results easily reproducible (the source code to conduct our experiments will also be made publicly available in an open-source software package).
First, we present experiments for object detectors trained on single categories
in Section~\ref{sec:single}---that is, detectors are trained individually for
each object category, and an experiment for the standard
multiple-category setting is presented in Section~\ref{sec:multiple}.
Finally, we present an ablation study in Section~\ref{sec:prelim} whose purpose
is to understand the effect of various factors (importance blending strategy,
placement strategy, and amount of labeled data).

\subsection{Dataset, Tools, and Metrics} \label{sec:datasets}

\paragraph{Dataset.}
In all our experiments, we use a subset of the Pascal VOC'12 training dataset
\cite{pascal} that contains segmentation annotations to train all our models
(context-model and object detector). We call this training set
\texttt{VOC12train-seg}, which contains $1\,464$ images. Following standard
practice, we use the test set of VOC'07 to evaluate the models, which contains
$4\,952$ images with the same 20 object categories as VOC'12. We call this image
set \texttt{VOC07-test}.

\paragraph{Object detector.}
To test our data-augmentation strategy we chose one of the state-of-the art
object detectors with open-source implementation, BlitzNet \cite{blitznet} that
achieves $79.1\%$ mAP on \texttt{VOC07-test} when trained on the union of the
full training and validation parts of VOC'07 and VOC'12, namely
\texttt{VOC07-train+val} and \texttt{VOC12train+val} (see~\cite{blitznet}); this
network is similar to the DSSD detector of \cite{dssd} that was also used in the
Focal Loss paper \cite{focal_loss}. The advantage of such class of detectors is
that it is relatively fast (it may work in real time) and supports training with
big batches of images without further modification.

\paragraph{Evaluation metric.}
In VOC'07, a bounding box is considered to be correct if its
Intersection over Union (IoU) with a ground truth box is higher than 0.5. The metric
for evaluating the quality of detection for one object class is the average precision (AP),
and the mean average precision (mAP) for the dataset.

\subsection{Implementation Details} \label{sec:details}

\paragraph{Selecting and blending objects.}
Since we widely use object instances extracted from the training images in all
our experiments, we create a database of objects cut out from the \texttt{VOC12train-seg}
set to quickly access them during training. For a given candidate
box, an instance is considered as matching if after scaling it by a factor in
$[0.5, 1.5]$ the re-scaled instance's bounding box fits inside the candidate's
one and takes at least 80\% of it’s area.
When blending them into the new background, we follow~\cite{cut_paste} and use
randomly one of the following methods: adding Gaussian or linear blur on the
object boundaries, generating blur on the whole image by imitating motion, or
just paste an image with no blending. To not introduce scaling artifacts, we
keep the scaling factor close to 1.

\begin{figure}[hbtp!]
\begin{center}
  \includegraphics[width=0.99\linewidth,trim=0 70 0 0,clip]{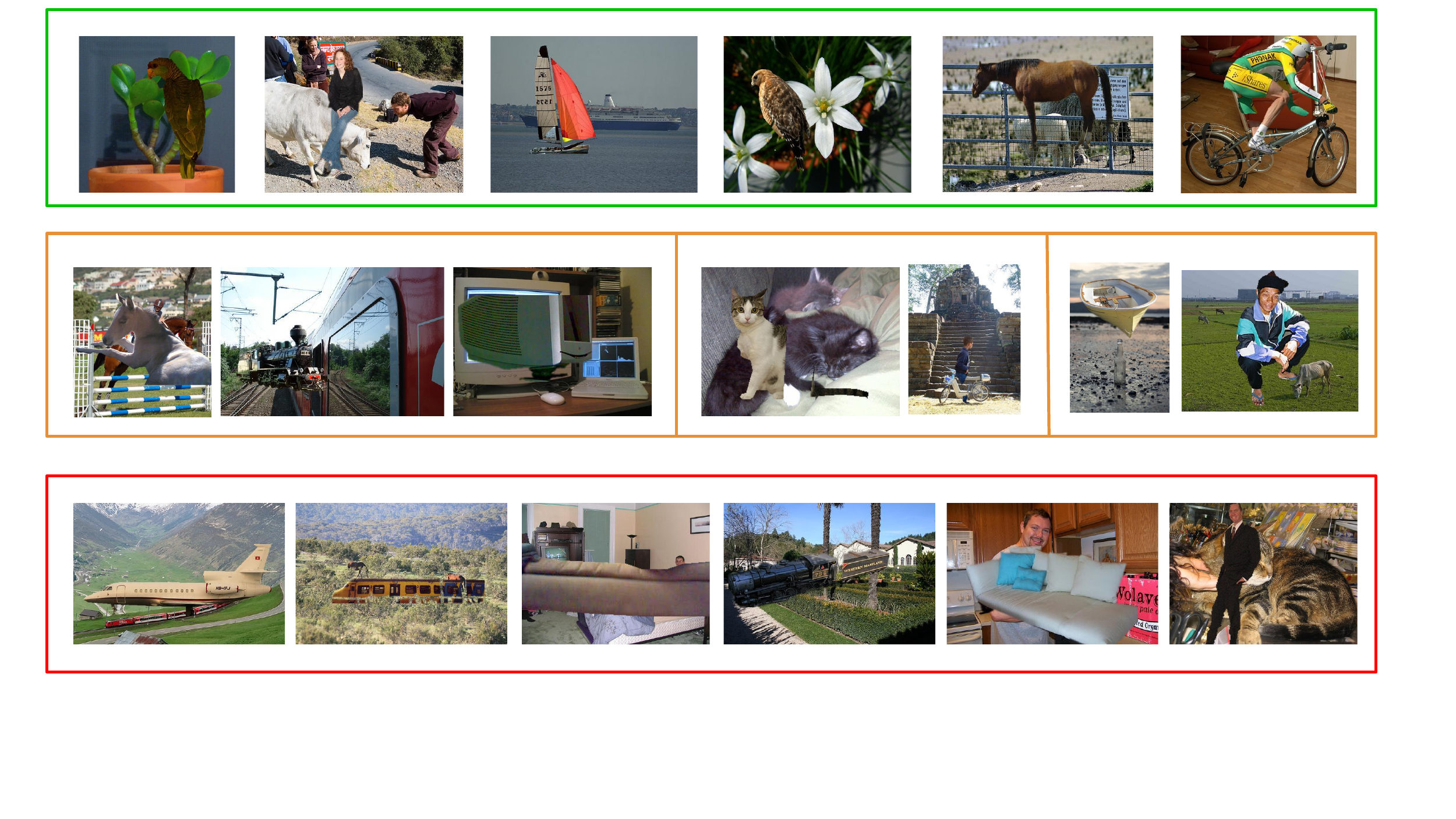}
\end{center}
\vspace*{-0.4cm}
\caption{\textbf{Examples of instance placement with context model guidance.}
  The figure presents samples obtained by placing a matched examples into the
   box predicted by the context model. The top row shows generated images that are
   visually almost indistinguishable from the real ones. The middle row presents
   samples of good quality although with some visual artifacts. For the two
   leftmost examples, the context module proposed an appropriate object class, but
   the pasted instances do not look visually appealing. Sometimes, the scene does
   not look natural because of the segmentation artifacts as in the two middle
  images. The two rightmost examples show examples where the category seems to be
  in the right environment, but not perfectly placed.
  The bottom row presents some failure cases.}
\label{fig:context_output}
\end{figure}

\paragraph{Training the context model.} After preparing the ``contextual images'' as
described in \ref{sec:context_inputs}, we re-scale them to the standard size
$300 \times 300$ and stack them in batches of size 32. We use ResNet50 \cite{resnet} with
ImageNet initialization to train a contextual model in all our experiments.
Since we have access only to the training set at any moment we train and
apply the model on the same data. To prevent overfitting, we use
early stopping. In order to determine when to stop the training procedure,
we monitor both training error on our training set and validation error on
the VOC'12 validation set \texttt{VOC12-val}. The moment when the loss curves
start diverging noticeably is used as a stopping point. To this end, when
building context model for one class
vs. background, we train a network for 1.5K iterations, then decrease the
learning rate by a factor 10 and train for 500 additional iterations. When
learning a joint contextual model for all 20 categories, we first run the
training procedure for 4K iterations and then for 2K more iterations after
decreasing the learning rate.
We sample 3 times more background contextual images, as noted in
Section~\ref{sec:context_inputs}. Visual examples of images produced by the
context model are presented in Figure~\ref{fig:context_output}. Overall,
training the context model is about 5 times faster than training the detector.

\paragraph{Training the object detector.}
In this work, the detector takes images of size $300 \times 300$ as an input and
produces a set of candidate object boxes with classification scores; like our
context model, it uses ResNet50 \cite{resnet} pre-trained on ImageNet as a
backbone. 
The detector is trained by following~\cite{blitznet}, with the ADAM optimizer
\cite{adam} starting from learning rate $10^{-4}$ and decreasing it later
during training by a factor 10 (see Sections~\ref{sec:single}
and~\ref{sec:multiple} for the number of epochs used in each experiment).
In addition to our data augmentation approach obtained by copy-pasting objects,
all experiments also include classical data augmentation steps obtained by
random-cropping, flips, and color transformations, following~\cite{blitznet}.

\subsection{Single-Category Object Detection}\label{sec:single}
In this section, we conduct an experiment to better understand the effect of
the proposed data augmentation approach, dubbed ``Context-DA'' in the different tables, when compared to a baseline with random
object placement ``Random-DA'', and when compared to standard data augmentation techniques called ``Base-DA''.
The study is conducted in a single-category setting, where detectors are
trained independently for each object category, resulting in a relatively small
number of positive training examples per class. This allows us to evaluate
the importance of context when few labeled samples are available and see if 
conclusions drawn for a category easily generalize to other ones.

The baseline with random object placements on random backgrounds is conducted
in a similar fashion as our context-driven approach, by following the strategy
described in the previous section.
For each category, we treat all images with no
object from this category as background images,
and consider a collection of cut instances as discussed in Section~\ref{sec:datasets}.
During training, we augment a
negative (background) image with probability 0.5 by pasting up to two instances
on it, either at randomly selected locations (Random-DA), or using our context model in the selected bounding boxes with top scores (Context-DA). The instances are re-scaled by a random factor
in $[0.5, 2]$ and blended into an image using a randomly selected blending
method mentioned in Section~\ref{sec:datasets}. For all models, we train the object detection network for 6K
iterations and decrease the learning rate after 2K and 4K iterations by a
factor 10 each time. 
The results for this experiment are presented in Table~\ref{tab:single_cat}.

The conclusions are the following: random placement indeed hurts the
performance on average. Only the category bird seems to benefit significantly
from it, perhaps because birds tend to appear in various contexts in this
dataset and some categories significantly suffer from random placement such as
boat, table, and sheep.  Importantly, the visual context model always improve
upon the random placement one, on average by 5\%, and upon the baseline that uses
only classical data augmentation, on average by 4\%. Interestingly, we identify
categories for which visual context is crucial (aeroplane, bird, boat, bus, cat, cow, horse),
for which context-driven data augmentation brings more than 5\% improvement and some categories
that display no significant gain or losses (chair, table, persons, train), where the difference
with the baseline is less than 1\%.

\begin{table*}[hbtp] 
\caption{Comparison of detection accuracy on \texttt{VOC07-test} for the single-category experiment.
  The models are trained independently for each category, by using the $1\,464$ images from \texttt{VOC12train-seg}.
  The first row represents the baseline experiment that uses standard data augmentation techniques. The second row
  uses in addition copy-pasting of objects with random placements. The third row presents the results achieved by 
  our context-driven approach and the last row presents the improvement it brings over the baseline. The numbers represent average precision per
  class in \%. Large improvements over the baseline (greater than $5\%$) are in bold.
}\label{tab:single_cat}
\centering
\renewcommand{\arraystretch}{1.3}
\renewcommand{\tabcolsep}{0.5mm}
\resizebox{\linewidth}{!}{
\begin{tabular}{l|c c c c c c c c c c c c c c c c c c c c |c|}
method & aero & bike & bird & boat & bott. & bus & car & cat & chair & cow & table & dog & horse & mbike & pers. & plant & sheep & sofa & train & tv & avg.\\
  \hline
Base-DA              & 58.8 & 64.3 & 48.8 & 47.8 & 33.9 & 66.5 & 69.7 & 68.0  &  40.4 & 59.0  & 61.0       & 56.2 & 72.1 & 64.2 & 66.7      & 36.6 & 54.5 & 53.0  & 73.4  & 63.6 & 58.0\\
Random-DA  & 60.2 & 66.5 & 55.1 & 41.9 & 29.7 & 66.5 & 70.0   &  70.1 & 37.4 & 57.4 & 45.3 & 56.7     & 68.3 & 66.1 &  67.0 & 37.0  & 49.9 &  55.8      & 72.1  & 62.6 & 56.9\\
Context-DA & 67.0  & 68.6 & 60.0   & 53.3 & 38.8 & 73.3 & 72.4 & 74.3 &  39.7 & 64.3 & 61.4      & 60.3 & 77.6 & 69.0  & 67.3      & 38.6 & 56.2 & 56.9 & 74.4  & 66.8 & 62.0\\
  \hline
Impr. Cont. &{\bf 8.2} &4.3 & {\bf 11.2} & {\bf 5.5} &4.9 & {\bf 6.8} &2.7 & {\bf 6.3} & -0.7 & {\bf 5.3} &0.4 &4.1 &{\bf 5.5} &4.8 &0.6 &2.0  &1.7 &3.9 &1.0 &3.2 &4.0\\
\end{tabular}
}
\vspace*{-0.5cm}
\end{table*}

\vspace*{-0.5cm}

\begin{table*}[hbtp] 
\caption{Comparison of detection accuracy on \texttt{VOC07-test} for the multiple-category experiment.
  The model is trained on all categories at the same time, by using the $1\,464$ images from \texttt{VOC12train-seg}.
  The first row represents the baseline experiment that uses standard data
augmentation techniques. The second row uses also our context-driven data
augmentation. The numbers represent average precision per class in \%.
}
\label{table:multiple}
\centering
\renewcommand{\arraystretch}{1.3}
\renewcommand{\tabcolsep}{0.5mm}
\resizebox{\linewidth}{!}{
\begin{tabular}{l|c c c c c c c c c c c c c c c c c c c c | c |}
method  & aero & bike & bird & boat & bott. & bus & car & cat & chair & cow & table & dog & horse & mbike & pers. & plant & sheep & sofa & train & tv & avg.\\
  \hline
  Base-DA & 63.6 & 73.3 & 63.2 & 57.0 & 31.5 & 76.0 & 71.5 & 79.9 & 40.0 & 71.6 & 61.4 & 74.6 & 80.9 & 70.4 & 67.9 & 36.5 & 64.9 & 63.0 & 79.3 & 64.7 & 64.6\\
  Context-DA & 66.8 & 75.3 & 65.9 & 57.2 & 33.1 & 75.0 & 72.4 & 79.6 & 40.6 & 73.9 & 63.7 & 77.1 & 81.4 & 71.8 & 68.1 & 37.9 & 67.6 & 64.7 & 81.2 & 65.5 & 65.9 
\end{tabular}
}
\vspace*{-0.5cm}
\end{table*}

\subsection{Multiple-Categories Object Detection}\label{sec:multiple}
In this section, we conduct the same experiment as in~\ref{sec:single}, but we train 
a single multiple-category object detector instead of independent ones per category.
Network parameters are trained with more labeled data (on average 20 times more
than for the models learned in Table~\ref{sec:single}), making them more robust to overfitting.
The results are presented in Table~\ref{table:multiple} and show a modest improvement
of $1.3\%$ on average over the baseline, which is relatively consistent across
categories, with 18 categories out of 20 that benefit from the context-driven
data augmentation. This confirms that data augmentation is mostly crucial when few labeled
examples are available.

\subsection{Ablation Study}\label{sec:prelim}
Finally, we conduct an ablation study to better understand 
(i) the importance of visual context for object detection,
(ii) the impact of blending artefacts, and (iii) the importance of data
augmentation when using very few labeled examples.
For simplicity, we choose the first 5 categories of VOC'12, namely \textit{aeroplane,
bike, bird, boat, bottle}, and train independent detectors per category as in
Section~\ref{sec:single}, which corresponds to a setting where few
samples are available for training.

\paragraph{Baseline when no object is in context.} 
Our experiments show that augmenting naively datasets with randomly placed objects slightly hurts the performance.
To confirm this finding, we consider a similar experiment, by learning
on the same number of instances as in Section~\ref{sec:single}, but we consider
as positive examples only objects that have been synthetically placed in a
random context. This is achieved by removing from the training data all the
images that have an object from the category we want to model, and replacing it
by an instance of this object placed on a background image.
The main motivation for such study is to consider the extreme case where (i)
no object is placed in the right context; (ii) all objects may suffer from
rendering artefacts. As shown in Table~\ref{tab:preliminary}, the average
precision degrades significantly by about $14\%$ compared to the baseline. As a
conclusion, either visual context is indeed crucial for learning, or blending
artefacts is also a critical issue. The purpose of the next experiment is to
clarify this ambiguity.

\begin{figure}[hbtp!]
\begin{center}
  \includegraphics[width=0.99\linewidth,trim=50 170 50 10,clip]{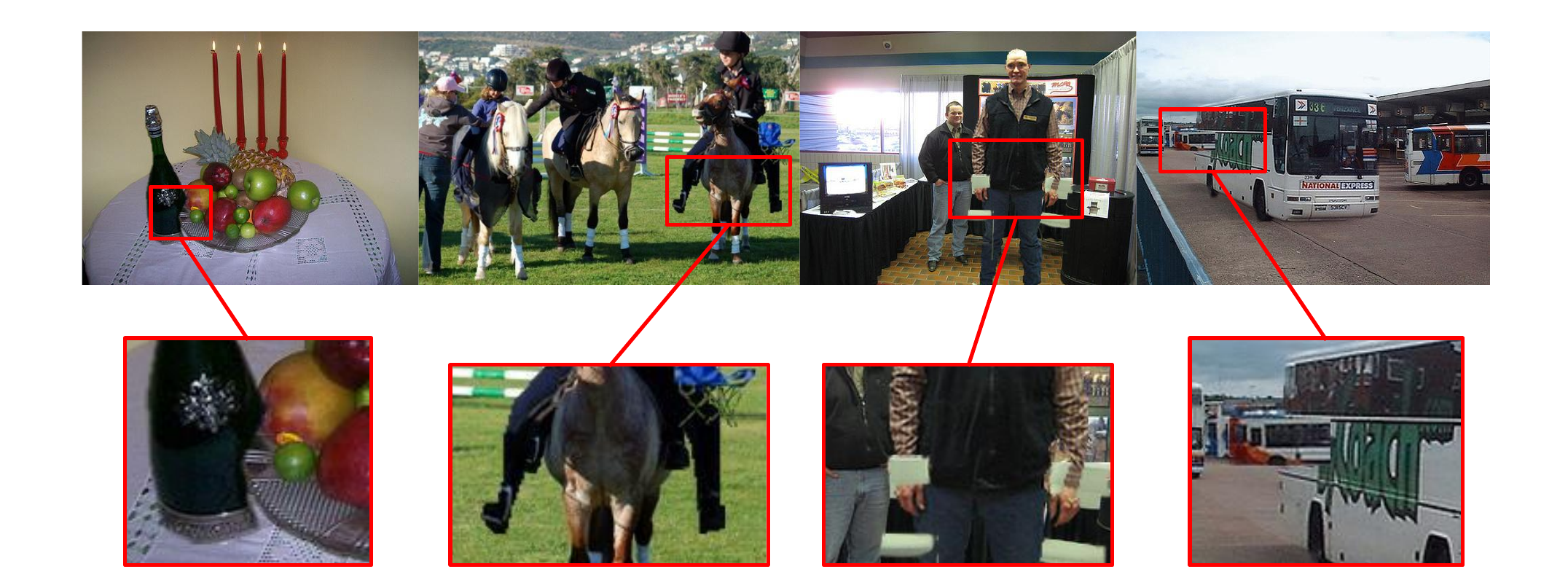}
\end{center}
\vspace*{-0.4cm}
\caption{\textbf{Illustration of artifacts arising from enlargement augmentation.}
  In the enlargement data augmentation, an instance is cut out of the image,
  up-scaled by a small factor and placed back at the same location. This approach 
  leads to blending artefacts.  Modified images are given in the top row.
  Zoomed parts of the images centered on blending artifacts are presented in the
  bottom line.}
\label{fig:enlargement}
\end{figure}

\paragraph{Impact of blending when the context is right.} 
In the previous experiment, we have shown that the lack of visual context and
the presence of blending artefacts may explain the performance drop observed
on the fourth row of Table~\ref{tab:preliminary}. Here, we propose a simple experiment
showing that blending artefacts are not critical when objects are placed in the
right context: the experiment consists of extracting 
each object instance from the dataset, up-scale it by a random factor slightly greater than one (in the interval $[1.2,1.5]$),
and blend it back at the same location, such that it covers the original
instance.
As a result, the new dataset benefits slightly from data augmentation
(thanks to object enlargement), but it also suffers from blending artefacts for
\emph{all object instances}. 
As shown on the fifth row of
Table~\ref{tab:preliminary}, this approach improves over
the baseline, though not as much as the full context-driven data augmentation,
which suggests that the lack of visual context was the key explaining the result
observed before. The experiment also confirms that the presence of
blending artefacts is not critical for the object detection task. Visual
examples of such artefacts are presented in Figure~\ref{fig:enlargement}.

\paragraph{Performance with very few labeled data.} 
Finally, the last four rows of Table~\ref{tab:preliminary} present the result
obtained by our approach when reducing the amount of labeled data, in a setting
where this amount is already small when using all training data. The improvement
provided by our approach is significant and consistent (about 6\% when using
only 50\% and 25\% of the training data). Even though one may naturally expect
larger improvements when a very small number of training examples are available,
it should be noted that in such very small regimes, the quality of the context
model may degrade as well (e.g., the dataset contains only 87 images of birds,
meaning that with 25\%, we use only 22 images with positive instances, which is
an extremely small sample size).
\vspace*{-0.3cm}

\begin{table}[hbtp!] 
\caption{Ablation study on the first five categories of VOC'12. All models are learned independently as in Table~\ref{tab:single_cat}. We compare classical data augmentation techniques (Base-DA), approaches obtained by copy-pasting objects, either randomly (Random-DA) or according to a context model (Context-DA). The line ``Removing context'' corresponds to the first experiment described in Section~\ref{sec:prelim}; Enlarge-Reblend corresponds to the second experiment, and the last four rows compare the performance of Base-DA and Context-DA when varying the amount of training data from 50\% to 25\%.
}
\label{tab:preliminary}
\centering
\begin{tabular}{l| c c c c c |c}
  Data portion & aero & bike & bird & boat & bottle & average \\
  \hline
  Base-DA             & 58.8 & 64.3 & 48.8 & 47.8 & 33.9 & 48.7\\
  Random-DA  & 60.2 & 66.5 & 55.1 & 41.9 & 29.7 & 48.3\\
  Context-DA  & 67.0 & 68.6 & 60.0 & 53.3 & 38.8 & 57.5 \\
  \hline
  Removing context & 44.0 & 46.8 & 42.0 & 20.9 & 15.5 & 33.9\\
  \hline
  Enlarge + Reblend-DA & 60.1 & 63.4 & 51.6 & 48.0 & 34.8 & 51.6\\
  \hline
  Base-DA 50 \%      & 55.6 & 60.1 & 47.6 & 40.1 & 21.0 & 42.2 \\
  Context-DA 50 \% & 62.2 & 65.9 & 55.2 & 46.9 & 27.2 & 48.8 \\
  Base-DA 25 \%      & 51.3 & 54.0 & 33.8 & 28.2 & 14.0 & 32.5 \\
  Context-DA 25 \%  & 57.8 & 59.5 & 40.6 & 34.3 & 19.0 & 38.3 \\
\end{tabular}
\vspace*{-0.5cm}
\end{table}


\section{Discussions and Future Work}
In this paper, we introduce a data augmentation technique dedicated to object
detection, which exploits segmentation annotations. From a methodological
point of view, we show that this approach is effective and goes beyond
traditional augmentation approaches.
One of the keys to obtain significant improvements in terms of accuracy was to
introduce an appropriate context model which allows us to automatically find
realistic locations for objects, which can then be pasted and blended at
in the new scenes. While the role of explicit context modeling has been
unclear so far for object detection, we show that it is in fact crucial when
performing data augmentation and learn with few labeled data, which is
one of the major issue that deep learning models are facing today.

We believe that these promising results pave the way to numerous extensions.
In future work, we will for instance study the application of our approach to other
scene understanding tasks, e.g., semantic or instance segmentation, and
investigate how to adapt it to larger datasets. Since our approach relies on 
pre-segmented objects, which are subsequently used for data augmentation, 
we are also planning to exploit 
automatic segmentation tools such as~\cite{liao2012building} in order to use
our method when only bounding box annotations are available.

\vspace*{0.3cm}
\noindent \textbf{Acknowledgments.} 
This work was supported by a grant from ANR (MACARON project
under grant number ANR-14-CE23-0003-01), by the ERC grant number 714381
(SOLARIS project), the ERC advanced grant ALLEGRO and gifts from Amazon and
Intel.

\clearpage

\bibliographystyle{splncs}
\bibliography{egbib}
\end{document}